\documentclass[letterpaper, 10 pt, conference]{ieeeconf}  

\IEEEoverridecommandlockouts                              

\usepackage{epsfig} 
\usepackage{mathptmx} 
\usepackage{amsmath} 
\usepackage{graphicx}
\usepackage{multirow}
\usepackage[ruled,linesnumbered]{algorithm2e}
\usepackage[bookmarks=true]{hyperref}

\title{\LARGE \bf
In-Hand Re-grasp Manipulation with Passive Dynamic Actions via Imitation Learning
}

\author{Dehao Wei$^{1}$, Guokang Sun$^{2}$, Zeyu Ren$^{2}$, Shuang Li$^{2}$, Zhufeng Shao$^{3}$,\\ Xiang Li$^{4}$, Nikos Tsagarakis$^{5}$ and Shaohua Ma$^{*}$      
\thanks{$^{1}$Dehao Wei is with Master Student of Tsinghua University, 100085, China.
        {\tt\small weidehao666@gmail.com}}%
\thanks{$^{3}$Zhufeng Shao and Xiang Li are with professor of Tsinghua University,
         515100, China.
        {\tt\small shaozf@mail.tsinghua.edu.cn, xiangli@tsinghua.edu.cn}}
\thanks{$^{4}$Nikos Tsagarakis is with professor of  Italian Institute of Technology(IIT),
         515100, Italy.
        {\tt\small nikos.tsagarakis@iit.it}}%
\thanks{$^{*}$Shaohua Ma is with professor of Tsinghua University,
        China, 100085, CN.
        {\tt\small mashaohua@sz.tsinghua.edu.cn}}
}
\begin{document}
\maketitle
\begin{figure*}[thpb]
  \centering
  \includegraphics[scale=0.98]{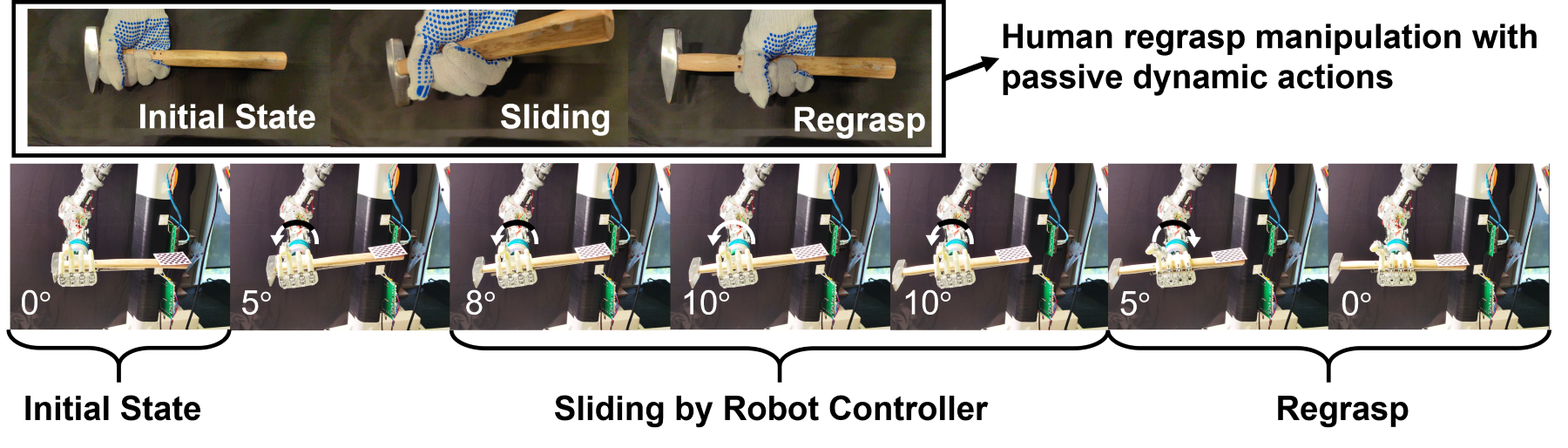}
  \caption{Overview of Manipulating a hammer by sliding using passive dynamic actions. The manipulation is shown from a side view. The human capability of re-grasping tools like a hammer using gravity as power source. In this study, we perform the hammer re-grasp task in three stages. The first stage represents the initial state, where the wrist of the robotic hand begins rotating without any sliding motion. In the second stage, the hammer slides within the hand due to gravity. Upon reaching the target position, the robotic hand is commanded to grasp the handle and return to the initial state. By utilizing our robot controller, we successfully identify the optimal gripping point on the hammer handle.}
  \label{figure1}
\end{figure*}

\thispagestyle{empty}
\pagestyle{empty}

\begin{abstract}
Re-grasp manipulation leverages on ergonomic tools to assist humans in accomplishing diverse tasks. In certain scenarios, humans often employ external forces to effortlessly and precisely re-grasp tools like a hammer. 
Previous development on controllers for in-grasp sliding motion using passive dynamic actions (e.g., gravity) relies on apprehension of finger-object contact information, and requires customized design for individual objects with varied geometry and weight distribution. It limits their adaptability to diverse objects.
In this paper, we propose an end-to-end sliding motion controller based on imitation learning~(IL) that necessitates minimal prior knowledge of object mechanics, relying solely on object position information. To expedite training convergence, we utilize a data glove to collect expert data trajectories and train the policy through Generative Adversarial Imitation Learning (GAIL). Simulation results demonstrate the controller's versatility in performing in-hand sliding tasks with objects of varying friction coefficients, geometric shapes, and masses.  
By migrating to a physical system using visual position estimation, the controller demonstrated an average success rate of 86$\%$, surpassing the baseline algorithm’s success rate of 35$\%$ of Behavior Cloning (BC) and 20$\%$ of Proximal Policy Optimization (PPO). More details are accessible through the link: \href{https://sites.google.com/view/dexmani/%E9%A6%96%E9%A1%B5}{Project website}
\end{abstract}

\section{INTRODUCTION}
Re-grasp manipulation plays a crucial role in utilizing human tools for various tasks. 
Humans possess the ability to skillfully and quickly re-grasp tools by leveraging external forces like gravity or interaction forces. This capability allows humans to achieve an optimal balance between precision and efficiency while performing tasks \cite{2}. Inspired by this characteristic of human manipulation, extensive research has focused on the utilization of external forces for in-hand manipulation. For instance, Nikhil Chavan-Dafle et al. explored extrinsic dexterity in re-grasp operations, such as transitioning from a power grasp to a precision grasp. They developed twelve hand-scripted re-grasp actions and evaluated their effectiveness through numerous trials involving different objects \cite{external_force}.

Instead of hand-scripted actions, Jian Shi et al. developed a framework for planning the motion of a robot hand to introduce an inertial load on a grasped object, enabling desired in-grasp sliding motion ~\cite{dynamic_in_hand_sliding_manipulation}. However, this framework lacks adaptability in unconditional or unknown object cases due to real-time feedback control limitations in finger motions and normal forces during re-grasping. Several studies have explored the concept of passive dynamic actions-based in-hand re-grasp operations. Erdmann and Mason employed this strategy to manipulate a planar object placed on a tray by controlling its tilt~\cite{doi:10.1177/027836499801700502}. Additionally, Erdmann demonstrated the similar technique for manipulating a 3D object with unknown geometry using two hands~\cite{Erdmann1986AnEO}. In both studies, a carefully planned sequence of contact mode transitions was executed, leveraging gravity to exert the required wrenching force on the manipulated object.
While several studies have developed dynamic motion controllers that effectively utilize external forces, particularly gravity, for manipulation, they are unsuitable for accommodating varying friction, mechanics models, or changes in geometry parameters and mass within the environment.
Deep Reinforcement Learing (DRL) method allows agents to optimize their strategies by continuously exploring the state action space~\cite{weidehao}. DRL has shown general capabilities in learning adaptive behaviors for complex and diverse scenarios such as dexterous manipulation and grasping~\cite{how_to_train_your_robot_with_drl,openai_learning_in_hand_dexterous_manipulation},  However, DRL methods typically take a long time to converge and require well-designed reward functions in order to learn simple target behaviors. 
GAIL is a imitation learning methods that take advantage of DRL, using a generative adversarial framework to learn policies from expert demonstrations~\cite{GAIL}, thus reducing the time needed to find improved control strategies.
GAIL offers robust and versatile imitation learning capabilities, handling unstructured data, transferring knowledge from limited demonstrations, and providing a balance between exploration and optimization. These benefits make it a valuable algorithm for various domains and tasks in the field of machine learning and robotics.
In this paper, we present the development of a robot controller utilizing imitation learning for the task of in-hand re-grasp manipulation, as illustrated in Figure \ref{figure1}. Our controller is characterized by its end-to-end design, which eliminates the dependence on friction and mechanics models. The main contributions of this study are summarized as follows:

\textbf{1)} We have proposed an initial end-to-end sliding motion controller utilizing imitation learning (IL), which demands minimal prior knowledge of object mechanics and relies exclusively on object position information. The controller has remarkable versatility in successfully executing in-hand sliding tasks involving objects with diverse friction coefficients, geometric shapes, and masses.

\textbf{2)} For the purpose of gathering expert data, We have developed an expert data collection platform that leverages data gloves to control robots within a simulation environment. Leveraging the acquired six expert trajectories, we trained the robot controller using Generative Adversarial Imitation Learning (GAIL). The GAIL algorithm exhibits convergence after approximately 400 episodes, while both BC and PPO algorithms have yet to converge even after the 500th episode.

\textbf{3)} We have devised an vision-based position estimation program, facilitating the transfer of our developed policy to real-world objects, thereby enhancing the practical applicability of our approach. In the real-world experiments of cylinder and cuboid, the average success rate achieved by the GAIL-based slide motion controller algorithm amounts to 0.86, outperforming BC’s 0.35 and PPO’s 0.20.

\begin{figure}[h]
  \centering
  \includegraphics[scale=0.25]{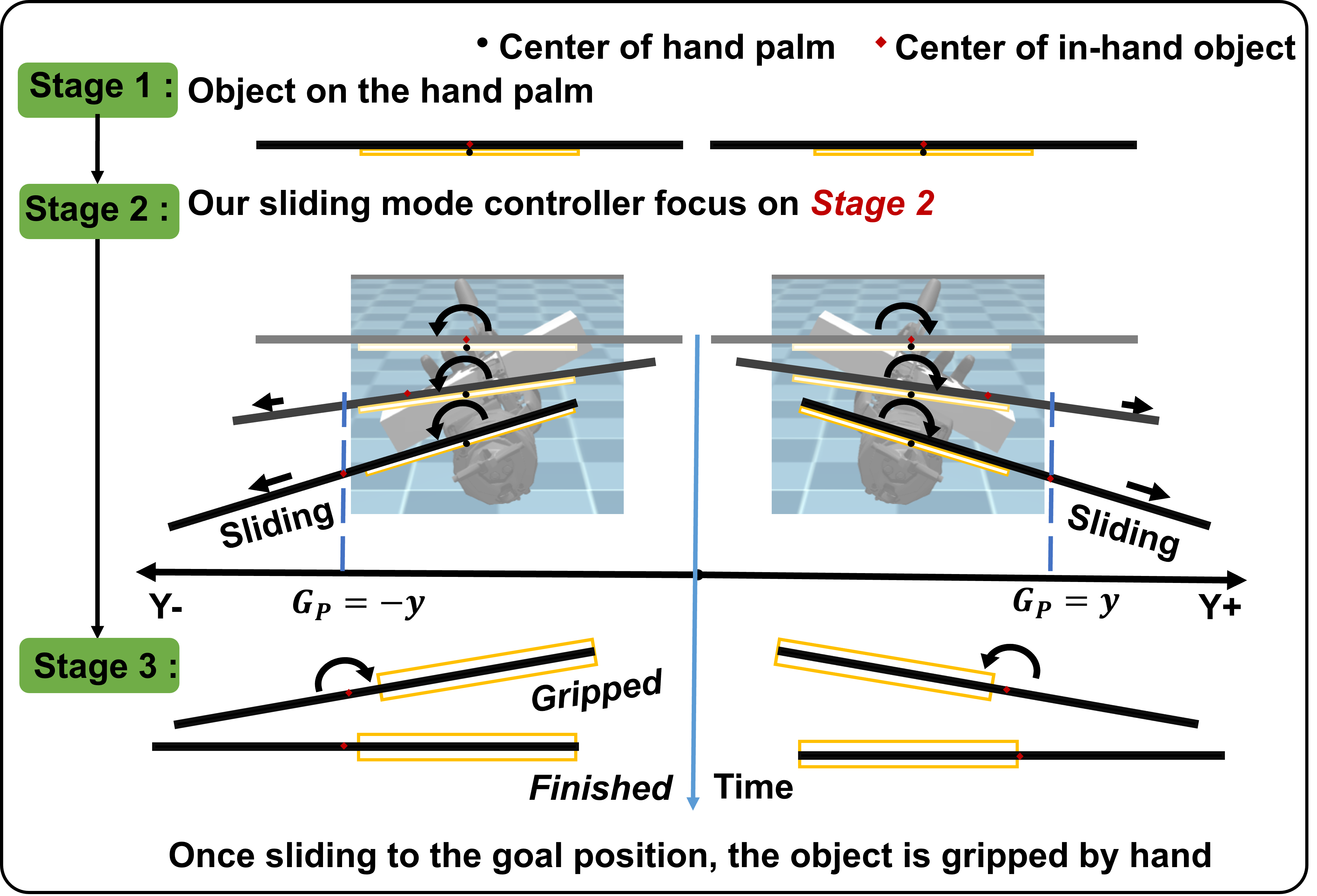}
  \caption{Overview of Task definition. The red diamond represents the center of mass (CM) of an object, while the black dot represents the center point of the robotic hand palm. In the left image, the target position is on the left side. As the mechanical hand rotates counterclockwise to a certain extent, the object starts sliding and reaches the desired position. The right image represents a similar situation where the target position is on the other side.}
  \label{figure2}
\end{figure}
\begin{figure*}[htbp]
  \centering
  \includegraphics[scale=0.63]{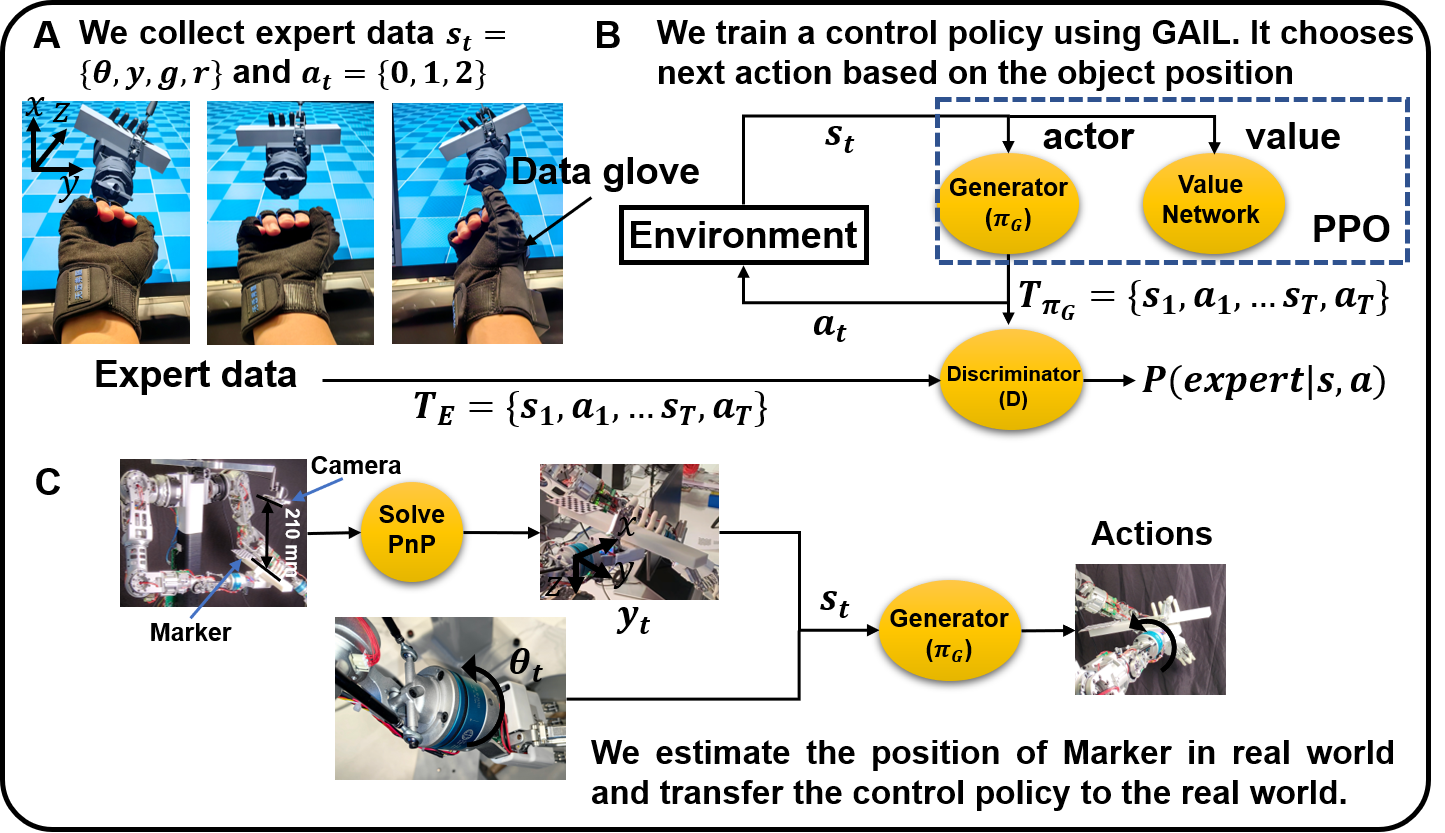}
  \caption{System overview. (a) Mapping relationship establishment between the data glove and the robot hand enables direct manipulation of the hand in the simulation environment, facilitating the capture of state-action pairs. (b) The collected expert data is utilized as input to train the Generator policy ($\pi_G$) using the GAIL trainer. The Discriminator, implemented as a three-layer neural network, discriminates between expert data ($T_E$) and the output of the Generator policy ($T_{\pi_G}$). Both the Discriminator and the Generator policy are three-layer neural networks. The training process of the Generator policy ($\pi_G$) is based on the PPO (Proximal Policy Optimization) algorithm, which serves as the baseline algorithm in this paper. (c) Transitioning to the real-world implementation, the object's position is determined by affixing a QR code onto the object. The Position Estimate algorithm is employed to extract the position information from the QR code. Additionally, the real-time feedback from the robot system is used to obtain the rotational position of the mechanical wrist. Combining the positional and rotational data, the control policy generates an action that is executed by the physical robot.}
  \label{figure3}
\end{figure*}
\begin{figure}[htbp]
  \centering
  \includegraphics[scale=0.42]{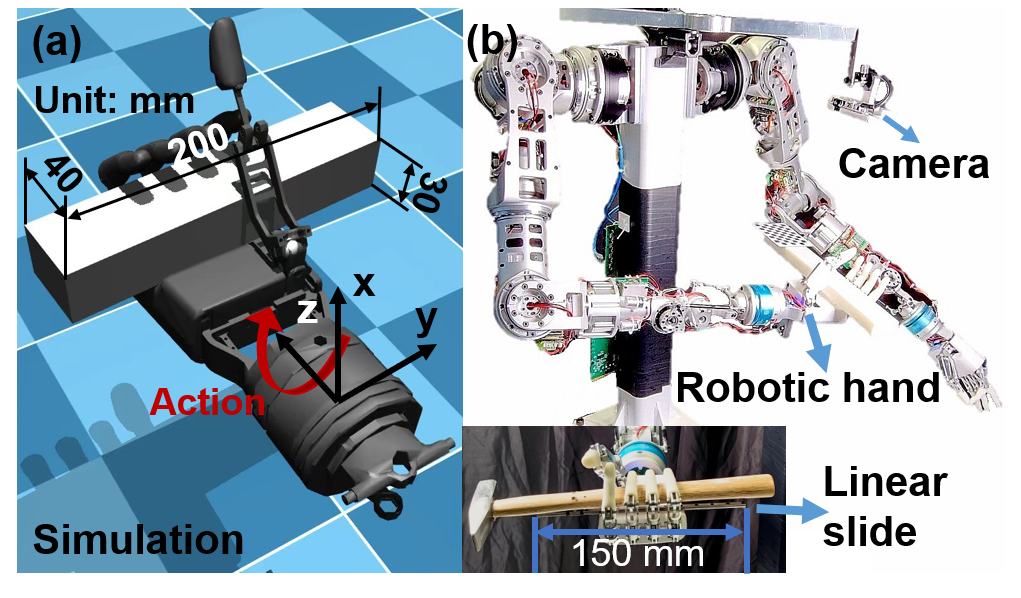}
  \caption{Simulation and Real World experiments setup}
  \label{figure4}
\end{figure}
\begin{figure*}[htbp]
  \centering
  \includegraphics[scale=0.48]{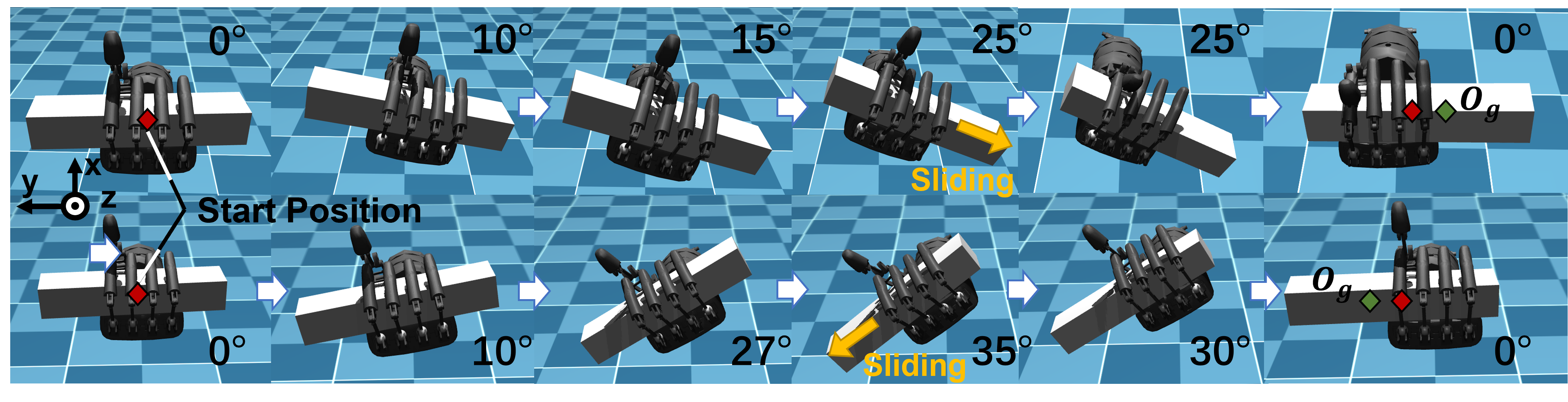}
  \caption{The demonstrations when the target points are on different side of y-axis in the simulation environment.}
  \label{figure5}
\end{figure*}

\section{RELATED WORK}
\subsection{In-hand Manipulation based on External Force}
Gravity is an ever-present power source and its applications have been explored for decades~\cite{external_force}. Francisco et al. proposed a sliding mode controller based on dynamic model for in-hand manipulation that re-positions a tool in the robot's hand by using gravity and controlling the slippage of the tool~\cite{in_hand_manipulation_using_gravity_and_controlled_slip}. Similar approaches are explored for planning and controlling in-hand manipulations by actively using gravity~\cite{vinaB} or dynamic motions ~\cite{holladay,Hou2016RobustPD,sintov}. Sundaralingam and Hermans proposed a purely kinematic approach for in-hand manipulation based on trajectory optimization with a multi-finger gripper. They assumed that the fingers on the object do not slip and impose soft constraints that encourage the minimization of finger slip. By assuming all finger contacts to be sticking, they bypassed the need for modeling the dynamics of contacts and obtained fast kinematic plans~\cite{sunda}. Subsequently in 2020, Nikhil Chavan-Dafle. et al. presented the mechanics and algorithms to compute the set of feasible motions of an object pushed in a plane. They generalized the motion cone construction to a broader set of planar tasks, such as those where external forces including gravity influence the dynamics of pushing. They showed that the motion cone was defined by a set of low-curvature surfaces and approximate it by a polyhedral cone~\cite{ijrr_planar_in_hand_manipulation_via_motion_cones}. These promising results on in-hand manipulation are limited to certain types of manipulation primitives or need to know the location and geometry of the features of object with precision and their friction coefficient. Demonstrating their applicability to general 3D manipulation with multiple contacts is largely missing since many of the assumptions made during their development get violated. For example, for a pushing case where a pusher force is not orthogonal to the normal force at support contacts, the development of motion cones in this thesis is no longer directly applicable. Motivated by this, in this study, we aim to develop a learning-based end-to-end slide  motion controller that need no primary knowledge of the mechanics or dynamic model.
\subsection{In-hand Manipulation with DRL and IL}
Deep reinforcement learning (DRL) provides a model-agnostic approach to control dynamical systems for manipulation. As a result there is a growing number of controller models based on deep reinforcement learning. Herry Zhu first extended Deep Reinforcement Learning-based dexterous manipulation to high-latitude robot control and achieved the execution of a number of dexterous manipulation actions~\cite{Zhu2018DexterousMW}.
However, reinforcement learning tends to require large data sets that are slow to converge, so IL methods are often used to speed up its training. 
In IL, demonstrations of expert are used to train policies that imitate the expert providing these successful trajectories. Standard approaches to learning from demonstrations are Behavior Cloning~(BC) and Inverse Reinforcement Learning~(IRL), which learns a policy through supervised learning to mimic the demonstrations. Although BC has been applied successfully in some instances like autonomous driving~\cite{bc}, it suffers from problems related to distribution drift~\cite{pmlr-v15-ross11a}. Furthermore, pure imitation learning methods cannot exceed the capabilities of the demonstrator since they lack a notion of task performance. An emerging derivative of IRL is represented by GAIL~\cite{GAIL} that learn how to act by directly learning the policy. GAIL has shown an empirical improvement in reducing the number of demonstrations required to successfully learn the task, compared to other imitation learning methods. Therefore we select GAIL as the method. In this study, state-action pairs are easily accessible through data gloves, so we can use state-action pairs from the entire expert trajectory for GAIL training. 

\section{Methods}
Our objective is to develop an optimal strategy that enables the re-grasp manipulation of in-hand objects. We initiate this task by collecting expert data, where a data glove is employed to instruct the robotic manipulator on how to grasp object poses. Subsequently, we employ the GAIL algorithm to train an optimal reinforcement learning strategy utilizing the acquired expert data.

\subsection{Task definition}\label{AA}
In this paper, we focus on the operation of adjusting the position of a tool in the hand, which is often encountered in decoration, parts assembly, and handicrafts. Specifically, we concentrate on tools with long handles such as hammers, as they facilitate gravity-based sliding, enabling the manipulator to rapidly re-grasp the hammer grip. During our training process, we utilize a cuboid object as the target for manipulation. This cuboid object has dimensions of 180 mm length, 40 mm width, and 30 mm height, with a mass of 156 g and a friction coefficient of 0.3.

In the subsequent variation stage, we test various objects, including cuboid, cylinders and hammers. This is possible because our end-to-end robot controller does not rely on precise geometric features and mechanical models. The parameters of these objects are listed in Table \ref{table1}. The object is placed on the robotic hand palm and restricted to move only along one degree of freedom, allowing movement along the y-axis. Figure \ref{figure2} illustrates our approach, where we use the geometric center of the object to represent its overall position, with the target position set on the y-axis.

The process consists of three stages. The Stage 1 is the initial state, in which the initial state is defined to be that the center of the object and the geometric center of the Robot hand are in the same perpendicular line. The robotic hand start to rotate around the Z-axis. In Stage 2, assuming that the robotic hand rotates by $\theta$, 

\begin{equation}
G*\sin\theta > f 
\end{equation}
where $G$ denotes the force of gravity and $f$ denotes the force of friction. Theoretically, the object will slowly slide by the distance of $d_{goal} $ and eventually reach the target position in the Y-axis. In Stage 3, upon reaching the target point, the manipulator will grip the object to prevent it from falling further, after which the manipulator returns to the initial position.
\begin{table}[h]
\caption{Physical properties of the experimental objects(D: Dimension, $F_C$: Friction coefficient).}
\label{table1}
\begin{center}
\begin{tabular}{|c|c|c|c|c|}
\hline
Shape & Material & D$[L,B,H]/[R,H] $ & Mass$(g)$ & $F_C$\\
\hline
Hammer Handle  & wood & 270,24,17 & 47.5 & 0.3\\
\hline
Hammer Head & steel  & 93,20,19 & 207.5 & 0.3\\
\hline
Cuboid & resinous & 320,40,20 & 161.3 & 0.3\\
\hline
Cylinder & resinous & 11,300 & 125.7 & 0.3\\
\hline
\end{tabular}
\end{center}
\end{table}
\subsection{Background of GAIL and DRL}
A general Reinforcement Learning (RL) problem can be formulated within the framework of a Markov Decision Process (MDP). In an MDP $M$, an agent learns by interacting with the environment. Specifically, $M$ is defined as a tuple $(S, A, r, P, \gamma, T)$, where $S$ represents the set of possible states, $A$ denotes the set of available actions, $r$ corresponds to the reward function, $P$ represents the transition probability distribution, $\gamma \in [0, 1]$ denotes the discount factor, and $T$ denotes the time horizon per episode.
In the RL setting, the environment produces a state observation $s_t \in S$ at each timestep $t$. Based on this observation, the agent samples an action $a_t$ from the policy distribution $\pi(s_t)$, where $a_t \in A$. The sampled action is then applied to the environment, causing the agent to transition to a new state $s_{t+1}$. The new state is sampled from the transition function $p(s_{t+1} | s_t, a_t)$, where $p \in P$. Alternatively, the episode may terminate at state $s_T$.
The primary objective of the RL agent is to learn a stochastic behavior policy $\pi_{\phi}: S \rightarrow P(A)$ parameterized by $\phi$. This stochastic policy aims to maximize the expected future discounted reward, denoted by $E[\sum_{i=0}^{T-1} \gamma^i r_i]$. Here, $\gamma$ is the discount factor, $r_i$ represents the reward obtained at timestep $i$, and $T$ denotes the time horizon per episode.
By formulating the RL problem within the MDP framework and defining the agent's goal, an effective approach to learning optimal decision-making policies in complex environments is provided.
\subsubsection{Proximal Policy Optimisation (PPO)}
PPO is an on-policy reinforcement learning algorithm that effectively handles both continuous and discrete action spaces \cite{ppo}. The PPO algorithm follows a two-step process, involving the collection of new observations and the improvement of the policy, while also approximating the value function \cite{ppo}. The update function for the PPO policy can be expressed as follows:
\begin{equation}
\scalebox{0.85}{$L(s_t, a_t, \theta_k, \theta) = \min \left( \frac{\pi_{\theta}(a_t|s_t)}{\pi_{\theta_k}(a_t|s_t)} \hat{A}^{\pi_{\theta_k}}(s_t, a_t), g(\epsilon, \hat{A}^{\pi_{\theta_k}}(s_t, a_t)) \right)$} 
\end{equation}
where $\theta_k$ are the parameters of the old policy $\pi_{\theta_k}$, and $g$ is defined as:

\begin{equation}
g(\epsilon, \hat{A}) = \begin{cases} 
(1 + \epsilon) \hat{A}_t,  \hat{A}_t \geq 0 \\
(1 - \epsilon) \hat{A}_t,  \hat{A}_t < 0 
\end{cases} 
\end{equation}

Here, $\hat{A}_t$ is the advantage estimator function at timestep $t$ and $\epsilon$ is a hyperparameter. The main concept underlying PPO is to ensure stable policy improvement by constraining the impact of policy updates through the utilization of the min operator.
\subsubsection{Generative Adversarial Imitation Learning (GAIL)}
GAIL, is an imitation learning algorithm \cite{GAIL} founded on generative adversarial networks (GANs) \cite{GAN}. GANs consist of a discriminator $D\phi'$ and a policy generator $G\phi$, with parameters denoted as $\phi$. The policy generator $G\phi$ produces exploration trajectories utilized by the discriminator to calculate a surrogate function, assessing the similarity between the generated policy and the expert policy. This similarity metric serves as a reward proxy for the reinforcement learning (RL) step. Unlike Inverse Reinforcement Learning (IRL) techniques, GAIL directly generates policies rather than the reward function.
The discriminator is trained to minimize the following loss function:
\begin{equation}
\footnotesize
L_{GAIL} = E_{\tau\phi}[\log(D\phi' (s_t, a_t))] + E_{\tau_E}[\log(1 - D\phi' (s_t, a_t))] 
\end{equation}
where $\tau\phi$ represents the trajectories generated by $G\phi$, and $\tau_E$ represents the expert trajectories.
The policy generator $G\phi$ is often adopted from methods based on stochastic policy, such as Proximal Policy Optimization (PPO) \cite{ppo}. PPO is the preferred choice in GAIL for two primary reasons: firstly, PPO employs a smooth policy update to ensure stable learning, and secondly, PPO generates diversified trajectories, expanding the sampling range available to the discriminator in GAIL.

\subsection{DRL Setup}
Standard DRL algorithms are trained in a simulation environment that is an exact replica of the real one. We consider PPO as the standard baseline DRL algorithm \cite{ppo}. 

\textbf{Reward:} In the training phase, we incentivize the object to approach the target position while imposing substantial penalties for instances where the object slips from the hand and descends to the ground. The reward function is defined as follows:
\begin{equation}
\text{Reward} = -\left(1-\Gamma\right) \left|\text{$O_p$} - \text{$G_p$}\right| - \lambda\Gamma
\end{equation}

Here, $\Gamma$ is a Boolean value,indicating whether the object has been dropped. $\lambda$ is the penalty coefficient, a critical hyperparameter governing the magnitude of penalties applied in cases where the object slips and descends to the ground($\lambda=10000$). $True$ indicates that the object has been dropped, and $False$ indicates that the object has not been dropped. $O_p$ represents the central position of the object. $G_p$ represents the goal position. $\Gamma$ is set to $True$ if Euclidean Distance between $O_p$ and $G_p$ is bigger than 200 $mm$.

\textbf{Action:} The action space is discrete and defined as $A \in \{0, 1, 2\}$. "0" represents "stop", "1" represents "step" forward by 0.01 radians, and "2" represents "step backward by 0.01 radians."

\textbf{State:} The state is defined as: $[W_p, O_p, G_p, R_p]$
where:$W_p$ represents the current rotation angle of the wrist joint. $R_p$ represents the absolute distance between $W_p$ and $C_p$. As shown in Fig. 2, the target position is set along the y-axis, with randomly assigned values of [-0.5, -0.3] and [0.3, 0.5](mm).

\textbf{Done:} The termination condition is defined as the distance bewteen $G_p$ and $W_p$ less than or equal to 10 mm.
\subsection{GAIL Setup}

To train a GAIL agent, a sequence of expert trajectories must be collected. In this study, as depicted in Fig. \ref{figure3}(a), expert data was gathered using a data glove (VRTRIX UE5). A mapping was established between the data glove and the robot. The motion information captured by the data glove was subsequently converted into control signals, enabling the control of the robotic arm in the simulation environment.
As generator agent within the GAIL framework, we employed the Proximal Policy Optimization (PPO) algorithm. For comparative purposes in future research, we also utilized the PPO algorithm and the Behavior Cloning algorithm as baselines. The neural network parameters are presented in Table \ref{Table2}, denoting the Neural Network (NN), Input Dimension (ID), Hidden Layer 1 Dimension (HL1), Hidden Layer 2 Dimension (HL2), Output Dimension (OD), and Activation Function (AF).

\begin{table}[h]
\caption{Parameters of the neutral networks used in this paper}
\label{Table2}
\begin{center}
\begin{tabular}{|c|c|c|c|c|c|}
\hline
NN & ID & HL1 & HL 2 & OD & AF\\
\hline
Generator($\pi_G$) & 4 & 128 & 128 & 4 & ReLU\\
\hline
Value Network & 4 & 128 & 128 & 3 & ReLU\\
\hline
Discrimantor(D) & 7 & 128 & 128 & 1 & ReLU\\
\hline
\end{tabular}
\end{center}
\end{table}

The generator agent $\pi_G$ and the expert data were discriminated using a discriminator, and the discriminator's output was utilized as the reward signal for updating the generator agent, as depicted in Fig. \ref{figure3}(b). Following the training process, the learned policy $\pi_G$ was transferred to a physical robot. To facilitate this transfer, as illustrated in Fig. \ref{figure3}(c), a marker(QR code) was affixed to determine the object's position. The Camera Calibration algorithm ~\cite{zhangzhengyou} was employ to acquire real-time positional information of the object within the hand. Furthermore, UDP communication was utilized to obtain real-time rotational pose information of the robotic arm's wrist. These measurements served as inputs to the policy $\pi_G$, with the output actions directly applied to the physical robot. Since the output values denoted a rotation increment of 0.01 radians per step, a PD algorithm was employed to control the rotational movement of the robotic arm's wrist($K_p= 0.1$).
\begin{figure*}[htbp]
  \centering
  \includegraphics[scale=0.46]{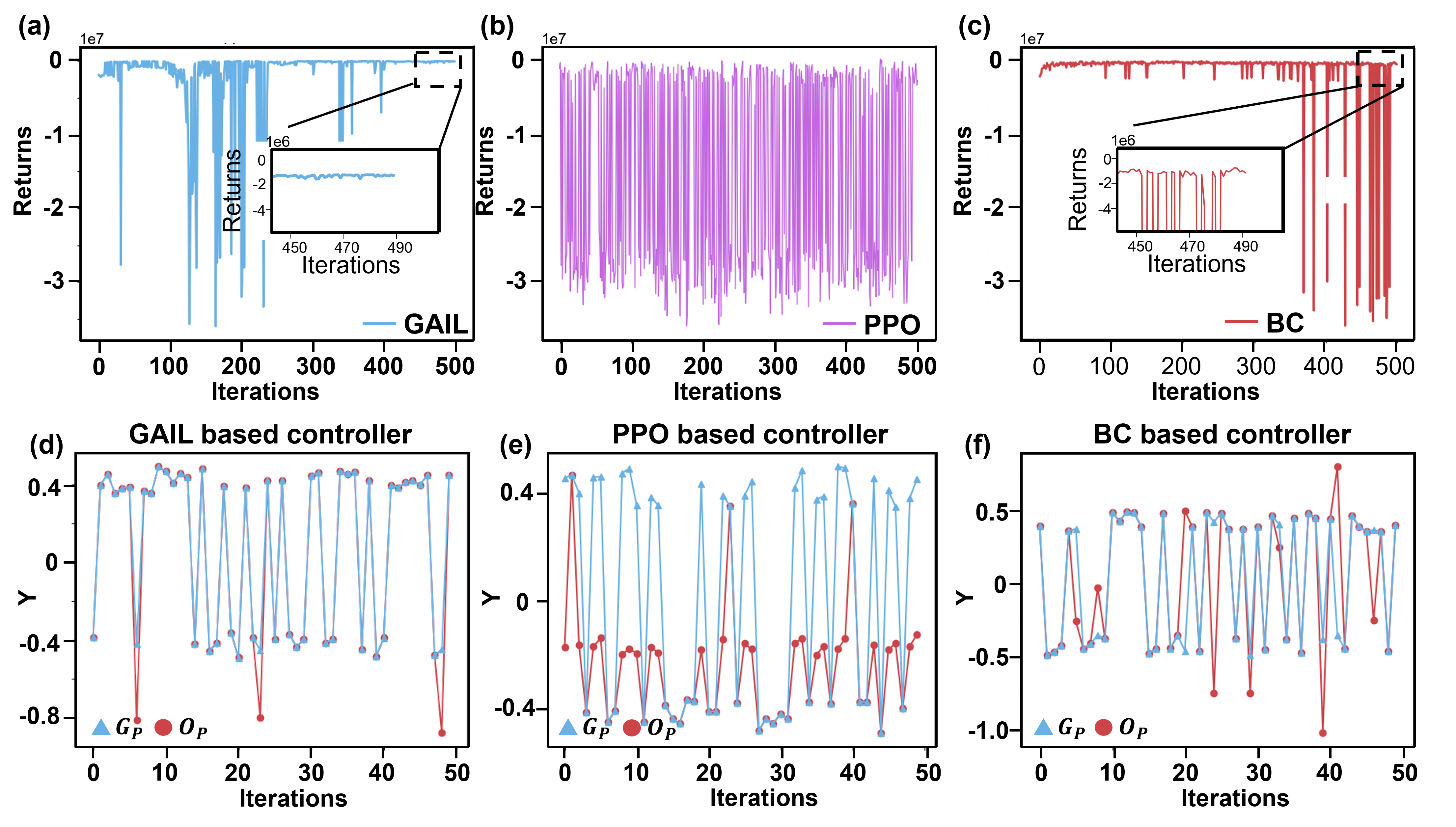}
  
  \caption{(a-c) Training curve of different algorithm (GAIL,PPO,BC). (b-d)Validation experiments of three algorithms conducted in a simulation environment. The figure displays the results from left to right, representing the BC, GAIL, and PPO algorithms. The experiment aimed to evaluate the manipulator's ability to adjust an object in its hand to a specified target position. The target position was a cuboid identical to the one used during the training process. The experiment was repeated 50 times for each algorithm. The success rates achieved by the three algorithms were as follows: GAIL - 0.94, BC - 0.82, and PPO - 0.52.$(G_p$:Goal Position, $O_p:$ Object Position$) $}
  \label{figure6}
\end{figure*}
\begin{figure*}[htbp]
  \centering
  \includegraphics[scale=1.24]{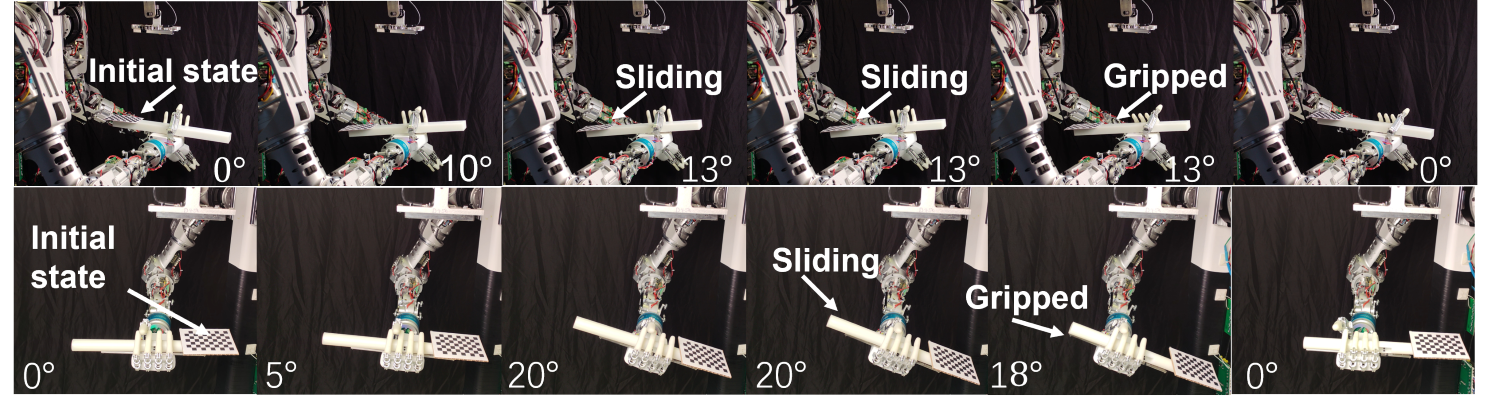}
  \caption{The demonstrations when the target points are on different side of y-axis in the Real world environment with the cuboid (above) and the cylinder (below). }
  \label{figure7}
\end{figure*}
\section{Experiment}

Our experiments encompass both simulation and real environments. Initially, our strategy is trained in the simulation environment, followed by its porting to the real machine environment. For the training task, we employ a cuboid as the manipulated object held in the hand. Subsequently, to assess the trained strategy's robustness, we apply it to re-gripping tasks involving a cylinder and a hammer tool. As shown in Fig. \ref{figure4}b, to ensure experimental safety and prevent potential injuries caused by the hammer's fall, we restrict the hammer's movement in the hand to a single degree of freedom along the Y-axis with the 150 mm linear slide.


\begin{table}[htbp]
\centering
\caption{Algorithm Parameters(NN: Neural Network)}
\label{teable3}
\begin{tabular}{|c|c|c|c|c|}
\hline
\textbf{Learning Rate} & \textbf{Discount Rate} & \textbf{Batch Size} & \textbf{Iterations} & \textbf{NN}\\
\hline
0.001 & 0.95 & 64 & 500 & FCN \\

\hline
\end{tabular}
\end{table}

\begin{table}[]
\centering
\caption{Success Rate in Simulation Environment}

\label{tab:my-table}
\begin{tabular}{|c|l|lll|lll|}
\hline
\multicolumn{1}{|l|}{} &                           & \multicolumn{3}{l|}{\textbf{CUBIOD}}                                                   & \multicolumn{3}{l|}{\textbf{CYLINDER}}                          \\ \hline
Algo                   & \multicolumn{1}{c|}{Mass(g)} & \multicolumn{1}{c|}{0.1}   & \multicolumn{1}{c|}{0.3} & \multicolumn{1}{c|}{0.5} & \multicolumn{1}{l|}{0.1} & \multicolumn{1}{l|}{0.3} & 0.5 \\ \hline
\multirow{3}{*}{\textbf{GAIL}}  & 120                      & \multicolumn{1}{l|}{  0.88}   & \multicolumn{1}{l|}{0.92}   &          0.86               & \multicolumn{1}{l|}{0.82}   & \multicolumn{1}{l|}{0.84}   &  0.84  \\ \cline{2-8} 
                       & 150                      & \multicolumn{1}{l|}{0.94}     & \multicolumn{1}{l|}{0.94}   &       0.96                  & \multicolumn{1}{l|}{0.88}   & \multicolumn{1}{l|}{0.86}   & 0.86   \\ \cline{2-8} 
                       & 200                      & \multicolumn{1}{l|}{0.88}     & \multicolumn{1}{l|}{0.84}  &        0.90                  & \multicolumn{1}{l|}{0.88}   & \multicolumn{1}{l|}{0.84}   &  0.91  \\ \hline

\multirow{3}{*}{\textbf{PPO}}   & 100                      & \multicolumn{1}{l|}{0.48} & \multicolumn{1}{l|}{0.46}   &       0.44                  & \multicolumn{1}{l|}{0.42}   & \multicolumn{1}{l|}{0.37}   &  0.46  \\ \cline{2-8} 
                       & 150                     & \multicolumn{1}{l|}{0.52} & \multicolumn{1}{l|}{0.56}  &           0.64              & \multicolumn{1}{l|}{0.52}   & \multicolumn{1}{l|}{0.54}   &  0.44  \\ \cline{2-8} 
                       & 200                      & \multicolumn{1}{l|}{0.53}     & \multicolumn{1}{l|}{0.54}   &       0.46                  & \multicolumn{1}{l|}{0.42}   & \multicolumn{1}{l|}{0.40}   &  0.46  \\ \hline
\multirow{3}{*}{\textbf{BC}}    & 100                      & \multicolumn{1}{l|}{0.66} & \multicolumn{1}{l|}{0.72}   &            0.78             & \multicolumn{1}{l|}{0.78}   & \multicolumn{1}{l|}{0.66}   &  0.72  \\ \cline{2-8} 
                       & 150                      & \multicolumn{1}{l|}{0.82}     & \multicolumn{1}{l|}{0.83}   &          0.81               & \multicolumn{1}{l|}{0.76}   & \multicolumn{1}{l|}{0.78}   &  0.76  \\ \cline{2-8} 
                       & 200                      & \multicolumn{1}{l|}{0.62}     & \multicolumn{1}{l|}{0.78}   &      0.80                  & \multicolumn{1}{l|}{0.68}   & \multicolumn{1}{l|}{0.64}   &  0.74  \\ \hline
\end{tabular}
\end{table}

\subsection{Simulation Experiment}

In the simulation experiment, depicted in Fig. \ref{figure4}a, we employed a 6-degree-of-freedom robotic hand and established its simulation environment using MuJoCo. 

Within this environment, the actions generated by the robot controller regulate the wrist's rotation around the Z-axis and the grasp state of the object. The wrist's rotation angle is constrained within $\pm$0.88 radians in simulation. The object's sliding within the hand is modeled as a variation in the Y-axis direction, considering only the change in the center position along the Y-axis as the state. Prior to reaching the target position, the manipulator remains in a slack phase without grasping the object, and the object is allowed to slide using passive dynamic action by our slide motion controller. Once it arrives at the target point $G_p$, the robotic hand securely grasps the object by prior hand programming. Since practical applications often require bi-directional adjustments of object position, we set the target positions as $G_p \in [-50,-30]$ or $[30,50]$ (mm). 

Fig. \ref{figure5} presents the demonstrations showcasing different target point locations with respect to the y-axis in the simulation environment. We trained a total of 500 episodes, where the termination condition for each episode is either reaching the goal or the object slipping out of the hand. In total, we trained three agents (GAIL, BC, PPO) where BC and PPO were used as baseline algorithms, as shown in Fig. \ref{figure6}(a)-(c). 
The three algorithms were trained under identical training environment, encompassing the same physical environment, initial conditions, and task specifications. Identical hyperparameter configurations were employed for the execution of each algorithm, encompassing parameters such as learning rate, discount factor, batch size, and neural network architecture, as shown in Table~\ref{teable3}. An equivalent number of training episodes were conducted for all algorithms, and testing was performed within the same environment for uniform evaluation. It can be seen that GAIL converges after 400 episodes of training, better than the Behavioral Cloning and PPO algorithms. We then validate the algorithm on the cuboid object in the simulation environment. 
As illustrated in Fig. \ref{figure6}(d)-(f), a series of 50 experiments were undertaken in which the robot controller was tasked to maneuver objects to randomized target positions. The experimental outcomes reveal that the robot controller, underpinned by the GAIL framework, achieved a commendable success rate of 0.94. This outperforms both the BC and PPO methodologies, which attained success rates of 0.82 and 0.54, respectively.
Additionally, to ascertain the broad applicability of our strategy, we modified the simulation experiments to encompass various friction coefficients (0.1, 0.3, and 0.5), object weights (100 g, 150 g, and 200 g), and object geometries. The selection of friction coefficients is determined by comparing them to the coefficients of friction found in a real-world linear guide environment. Subsequently, we carried out the experiments using two objects, namely the cylinder and the cuboid, both capable of sliding in the hand to facilitate position adjustment. The dimensions of the cuboid were kept consistent with the previous experiments, while the cylinder had a cross-sectional radius of 15 mm and a height of 200 mm. We conducted the experiments individually using GAIL, PPO, and BC controllers, and the corresponding results are presented in Table 4.
The GAIL controller demonstrated an average success rate of 93$\%$, surpassing the PPO and BC controllers, which achieved success rates of 54$\%$ and 78$\%$ respectively. Two primary factors contribute to the disparity in success rates. Firstly, excessive slipping of the object rendered the GAIL controller unsuccessful in the 2nd, 5th, and 29th experiments involving the cuboid object weighing 120 g and having a friction coefficient of 0.1. Secondly, the algorithm failed to converge within the 500 training episodes, leading to sub-optimal decisions by the BC and PPO controllers.
\subsection{Real Robotic Experiment}
\begin{table}[h]
\caption{Success rate in real environment}
\label{table5}
\begin{center}
\begin{tabular}{|c|c|c|}
\hline
\textbf{Algorithm} & \textbf{Cylinder} & \textbf{Cuboid}\\
\hline
GAIL & 0.88 & 0.84\\
\hline
PPO & 0.14 & 0.26\\
\hline
BC & 0.34 & 0.36\\
\hline
\end{tabular}
\end{center}
\end{table}
\begin{algorithm}
    \caption{Marker Position Estimation}
    \label{alg}
    \DontPrintSemicolon
    \SetKwData{Frame}{$f$}
    \SetKwData{CameraMatrix}{$c_m$}
    \SetKwData{DistCoeffs}{$d_c$}
    \SetKwData{QRCodes}{$q_r$}
    \SetKwData{QRDetector}{$q_d$}
    \SetKwData{ColorFrame}{$c_f$}
    \SetKwData{GrayscaleFrame}{$g_f$}
    \SetKwData{DecodedQRCode}{$d_q$}
    \SetKwData{currenttime}{$t$}
    \CameraMatrix $\leftarrow$ Camera matrix $p_c$,
    \DistCoeffs $\leftarrow$ Distortion coefficients $d_c$, \currenttime $\leftarrow$ Current period $t$,
    \QRDetector $\leftarrow$ Marker detector\;
    Start Realsense camera capture\;
    \While{True}{
        \Frame $\leftarrow$ Wait for a new frame from the camera\;
        \If{\Frame exists}{
            \ColorFrame $\leftarrow$ $f_t$\;
            Convert and Undistort \ColorFrame using \CameraMatrix and \DistCoeffs\;
            
            Convert \ColorFrame to grayscale\;
            \QRCodes $\leftarrow$ Detect Marker in \GrayscaleFrame using \QRDetector\;
            \If{\QRCodes are detected}{
                \ForEach{\DecodedQRCode in \QRCodes}{
                    Calculate the position of \DecodedQRCode in camera coordinates\;
                }
            }
        }
    }
\end{algorithm}
When conducting experiments on the real robot, precise acquisition of the object's pose proves to be paramount. here, a marker pose estimation algorithm based on the camera calibration method~\cite{zhangzhengyou} was developed, utilizing the Realsense 450 camera from Intel Company shown as Algorithm~\ref{alg}. 
In accordance with Figure~\ref{figure7}, the marker ($87 \times 62 mm$) is positioned atop the object. To mitigate potential obstruction of the marker during object sliding, it is strategically placed in proximity to the object's terminus.
Additionally, a guide rail was installed on the palm to constrain the object's freedom of movement. Various objects, including a hammer, cylinder, and cube, were utilized in the conducted experiments. The success rates for the experiments involving the cylinder and cube objects, as depicted in Figure \ref{figure7}. 
As shown in Table \ref{table5}, in the real-world experiments of cylinder and cuboid, the average success rate  achieved by the GAIL-based slide motion controller algorithm amounts to 0.86, outperforming BC's 0.35 and PPO's 0.20.

It should be noted that the success rate in the real experiments is considerably lower than that observed in the simulation experiments. This disparity can be attributed to several factors, including the restricted frame rate of the camera, communication delays, and the physical movement time of the robot. Consequently, the manipulator often failed to grasp the object in time while sliding towards the target position. 
Excessive object sliding within the grasp represents a significant factor contributing to the majority of failures observed in real-world experiments. This occurrence can be attributed to the challenging nature of swiftly reacting and grasping the object when it surpasses the intended target position due to the high velocity involved in the sliding process.

\section{CONCLUSIONS}
This paper presents an end-to-end in-hand re-grasp manipulation controller based on Generative Adversarial Imitation Learning (GAIL). The controller is designed to operate without necessitating finger-object contact data and adeptly transfers learned policies to objects characterized by distinct weights (120 g, 150 g, and 200 g), frictional coefficients (0.3, 0.4, and 0.5), and geometries (cuboid and cylinder). Such adaptability underscores the controller's extensive utility. In the proposed methodology, a expert data collection system is deployed that utilizes data gloves to dictate the movements of a robotic hand within a simulated environment, thereby enabling the acquisition of expert trajectory data. By utilizing only six expert trajectories, the GAIL controller demonstrates significantly faster convergence rates and robust performance compared to traditional baseline algorithms such as Behavioral Cloning (BC) and Proximal Policy Optimization (PPO) under the same training level. A vision-based position estimation system has been developed for real-time object position acquisition. Subsequently, the derived policy is implemented on real humanoid robots, successfully demonstrating multiple manipulation tasks. 

Overall, the proposed GAIL-based end-to-end controller for in-hand re-grasp manipulation offers improved robustness and convergence compared to traditional baseline algorithms. The utilization of expert trajectories collected through the data glove-based system contributes to the augmented performance of the controller. 
The future work will focus on addressing the issue of excessive sliding that prevents the robotic hand from grasping objects when they reach the target position. This problem can be tackled through the design of a faster reactive control system for the robotic hand. Additionally, the perception system can be further improved by employing higher-precision visual sensors to quickly and accurately determine the target position.

\addtolength{\textheight}{-12cm}   



\section*{ACKNOWLEDGMENT}

This work gets help from Yinghao Gao, Jiajun Zhang, Lu Qi, Liqun Huang, Yinpeng Qu and Chenguang Sun. This work is funded by National Natural Science Foundation of China, Joint Fund, grant numbers U19A20101, 82341019, and 61971255.
\bibliographystyle{ieeetr}
\bibliography{reference}

\begin{thebibliography}{10}

\bibitem{2}
A.~Okamura, N.~Smaby, and M.~Cutkosky, ``An overview of dexterous manipulation,'' in {\em Proceedings 2000 ICRA. Millennium Conference. IEEE International Conference on Robotics and Automation. Symposia Proceedings (Cat. No.00CH37065)}, vol.~1, pp.~255--262 vol.1, 2000.

\bibitem{external_force}
N.~C. Dafle, A.~Rodriguez, R.~Paolini, B.~Tang, S.~S. Srinivasa, M.~Erdmann, M.~T. Mason, I.~Lundberg, H.~Staab, and T.~Fuhlbrigge, ``Extrinsic dexterity: In-hand manipulation with external forces,'' in {\em 2014 IEEE International Conference on Robotics and Automation (ICRA)}, pp.~1578--1585, 2014.

\bibitem{dynamic_in_hand_sliding_manipulation}
J.~Shi, J.~Z. Woodruff, P.~B. Umbanhowar, and K.~M. Lynch, ``Dynamic in-hand sliding manipulation,'' {\em IEEE Transactions on Robotics}, vol.~33, no.~4, pp.~778--795, 2017.

\bibitem{doi:10.1177/027836499801700502}
M.~Erdmann, ``An exploration of nonprehensile two-palm manipulation,'' {\em The International Journal of Robotics Research}, vol.~17, no.~5, pp.~485--503, 1998.

\bibitem{Erdmann1986AnEO}
M.~A. Erdmann and M.~T. Mason, ``An exploration of sensorless manipulation,'' {\em Proceedings. 1986 IEEE International Conference on Robotics and Automation}, vol.~3, pp.~1569--1574, 1986.

\bibitem{weidehao}
D.~Wei, J.~Zhou, Y.~Zhu, and et~al., ``{Axis-space framework for cable-driven soft continuum robot control via reinforcement learning},'' {\em {Communication Engineering}}, vol.~2, p.~61, 2023.

\bibitem{how_to_train_your_robot_with_drl}
J.~Ibarz, J.~Tan, C.~Finn, M.~Kalakrishnan, P.~Pastor, and S.~Levine, ``How to train your robot with deep reinforcement learning: lessons we have learned,'' {\em The International Journal of Robotics Research}, vol.~40, no.~4-5, pp.~698--721, 2021.

\bibitem{openai_learning_in_hand_dexterous_manipulation}
O.~M. Andrychowicz, B.~Baker, M.~Chociej, R.~Józefowicz, B.~McGrew, J.~Pachocki, A.~Petron, M.~Plappert, G.~Powell, A.~Ray, J.~Schneider, S.~Sidor, J.~Tobin, P.~Welinder, L.~Weng, and W.~Zaremba, ``Learning dexterous in-hand manipulation,'' {\em The International Journal of Robotics Research}, vol.~39, no.~1, pp.~3--20, 2020.

\bibitem{GAIL}
J.~Ho and S.~Ermon, ``Generative adversarial imitation learning,'' in {\em Advances in Neural Information Processing Systems} (D.~Lee, M.~Sugiyama, U.~Luxburg, I.~Guyon, and R.~Garnett, eds.), vol.~29, Curran Associates, Inc., 2016.

\bibitem{in_hand_manipulation_using_gravity_and_controlled_slip}
F.~E. Viña~B., Y.~Karayiannidis, K.~Pauwels, C.~Smith, and D.~Kragic, ``In-hand manipulation using gravity and controlled slip,'' in {\em 2015 IEEE/RSJ International Conference on Intelligent Robots and Systems (IROS)}, pp.~5636--5641, 2015.

\bibitem{vinaB}
F.~E. Viña~B., Y.~Karayiannidis, C.~Smith, and D.~Kragic, ``Adaptive control for pivoting with visual and tactile feedback,'' in {\em 2016 IEEE International Conference on Robotics and Automation (ICRA)}, pp.~399--406, 2016.

\bibitem{holladay}
A.~Holladay, R.~Paolini, and M.~T. Mason, ``A general framework for open-loop pivoting,'' in {\em 2015 IEEE International Conference on Robotics and Automation (ICRA)}, pp.~3675--3681, 2015.

\bibitem{Hou2016RobustPD}
Y.~Hou, Z.~Jia, A.~M. Johnson, and M.~T. Mason, ``Robust planar dynamic pivoting by regulating inertial and grip forces,'' in {\em Workshop on the Algorithmic Foundations of Robotics}, 2016.

\bibitem{sintov}
A.~Sintov and A.~Shapiro, ``Swing-up regrasping algorithm using energy control,'' in {\em 2016 IEEE International Conference on Robotics and Automation (ICRA)}, pp.~4888--4893, 2016.

\bibitem{sunda}
B.~Sundaralingam and T.~Hermans, ``Relaxed-rigidity constraints: In-grasp manipulation using purely kinematic trajectory optimization,'' in {\em Robotics: Science and Systems}, 2017.

\bibitem{ijrr_planar_in_hand_manipulation_via_motion_cones}
N.~Chavan-Dafle, R.~Holladay, and A.~Rodriguez, ``Planar in-hand manipulation via motion cones,'' {\em The International Journal of Robotics Research}, vol.~39, no.~2-3, pp.~163--182, 2020.

\bibitem{Zhu2018DexterousMW}
H.~Zhu, A.~Gupta, A.~Rajeswaran, S.~Levine, and V.~Kumar, ``Dexterous manipulation with deep reinforcement learning: Efficient, general, and low-cost,'' {\em 2019 International Conference on Robotics and Automation (ICRA)}, pp.~3651--3657, 2018.

\bibitem{bc}
M.~Bojarski, D.~W. del Testa, D.~Dworakowski, B.~Firner, B.~Flepp, P.~Goyal, L.~D. Jackel, M.~Monfort, U.~Muller, J.~Zhang, X.~Zhang, J.~Zhao, and K.~Zieba, ``End to end learning for self-driving cars,'' {\em ArXiv}, vol.~abs/1604.07316, 2016.

\bibitem{pmlr-v15-ross11a}
S.~Ross, G.~Gordon, and D.~Bagnell, ``A reduction of imitation learning and structured prediction to no-regret online learning,'' in {\em Proceedings of the Fourteenth International Conference on Artificial Intelligence and Statistics} (G.~Gordon, D.~Dunson, and M.~Dudík, eds.), vol.~15 of {\em Proceedings of Machine Learning Research}, (Fort Lauderdale, FL, USA), pp.~627--635, PMLR, 11--13 Apr 2011.

\bibitem{ppo}
J.~Schulman, F.~Wolski, P.~Dhariwal, A.~Radford, and O.~Klimov, ``Proximal policy optimization algorithms,'' {\em ArXiv}, vol.~abs/1707.06347, 2017.

\bibitem{GAN}
I.~Goodfellow, J.~Pouget-Abadie, M.~Mirza, B.~Xu, D.~Warde-Farley, S.~Ozair, A.~Courville, and Y.~Bengio, ``Generative adversarial networks,'' {\em Commun. ACM}, vol.~63, p.~139–144, oct 2020.

\bibitem{zhangzhengyou}
Z.~Zhang, ``A flexible new technique for camera calibration,'' {\em IEEE Transactions on Pattern Analysis and Machine Intelligence}, vol.~22, no.~11, pp.~1330--1334, 2000.

\end{thebibliography}
\end{document}